# Advanced Hough-based method for on-device document localization

*D.V. Tropin [1,2,5], A.M. Ershov [3,5], D.P. Nikolaev [4,5], V.V. Arlazarov [2,5]*
[1] *Moscow Institute of Physics and Technology (National Research University), Dolgoprudny, Russia,*
[2] *FRC CSC RAS, Moscow, Russia,*
[3] *Moscow State University, Moscow, Russia,*
[4] *Institute for Information Transmission Problems of the RAS (Kharkevich Institute), Moscow, Russia,*
[5] *LLC "Smart Engines Service", Moscow, Russia.*

*Abstract*

The demand for on-device document recognition systems increases in conjunction with the emergence of more strict privacy and security requirements. In such systems, there is no data transfer from the end device to a third-party information processing servers. The response time is vital to the user experience of on-device document recognition. Combined with the unavailability of discrete GPUs, powerful CPUs, or a large RAM capacity on consumer-grade end devices such as smartphones, the time limitations put significant constraints on the computational complexity of the applied algorithms for on-device execution.

In this work, we consider document location in an image without prior knowledge of the document content or its internal structure. In accordance with the published works, at least 5 systems offer solutions for on-device document location. All these systems use a location method which can be considered Hough-based. The precision of such systems seems to be lower than that of the state-of-the-art solutions which were not designed to account for the limited computational resources.

We propose an advanced Hough-based method. In contrast with other approaches, it accounts for the geometric invariants of the central projection model and combines both edge and color features for document boundary detection. The proposed method allowed for the second best result for SmartDoc dataset in terms of precision, surpassed by U-net like neural network. When evaluated on a more challenging MIDV-500 dataset, the proposed algorithm guaranteed the best precision compared to published methods. Our method retained the applicability to on-device computations.

<u>Key words</u>: document detection, rectangle object localization, smartphone-based acquisition, on-device recognition, Hough transform, image segmentation

<u>Citation</u>: Tropin DV, Ershov AM, Nikolaev DP, Arlazarov VV. Advanced Hough-based method for on-device document localization. Computer Optics 20XX; 4X(X): XXX-YYY. DOI: 10.18287/2412-6179-CO-editorial index.

<u>Acknowledgments</u>: This work is partially supported by Russian Foundation for Basic Research (projects 18-29-26035 and 19-29-09092).

*Introduction*

Systems for document recognition [1, 2] on the image have been successfully used for a long time in various institutions and industries such as customs, transport, banking, trade, and telecommunications. With a significant development of Privacy and Security Standards [3], the demand for on-device document recognition systems which do not transfer information to third-party recognition servers is also increasing. Such systems are used for remote authentication in car-sharing rentals, for automating bank card entry for online shopping, and for business card scanning. To improve the user's experience when interacting with such systems, it is necessary to reduce the response time of the system. In the general case, when a mobile device has a weak CPU, little RAM, and does not have a GPU, the time constraints imply the employment of algorithms with low workloads.

In this paper, we consider the localization of a document in an image. The problem statement is as follows: let the document be a rigid flat rectangle with a known aspect ratio and an unknown content. The image is captured with a calibrated camera with a known back focal length and a principle point in the center of the image. The resulting frame contains an image of only one document with a given aspect ratio. The slop of the two facing sides of the document is within $(-1;1]$ for each (let us call them primarily horizontal), and the slop of the other pair of opposite sides is within $(-\infty,1] \cup (1,+\infty)$ for each (primarily vertical). One side of the document may be partially or completely obscured, out of the frame, or have low contrast, which is a common case of hand-held photography. The task is to define the quadrilateral formed by the outer borders of the document up to the vertex renumbering (the problem of document recognition, which is beyond the scope of this article, could be solved using method invariant to 180 degree rotation angle [4]). The typical inherent limitations of a smartphone can be summarized as follows. The average response time is 100 ms per frame on the Apple A8 processor (iPhone 6) in a single-threaded mode.

Algorithms for localization of documents with unknown content can be divided into 3 classes: based on the document borders detection (we will refer to them as Hough-based since the Hough transform is used to find the document borders), based on the salient region detection, and based on the document vertices detection.

The Hough-based algorithm was originally employed in [5]. In this work, Zhang and He proposed the following pipeline: borders detection, search for straight-lines using the Hough transform, formation, filtering, and ranking of quadrilaterals, refinement of the quadrilateral with the highest score. Authors consider the whiteboard localization. Thus, they imply that boundaries of the whiteboard are visible: the major part of each side is within the frame, has strong contrast, and not occluded. For the localization of the hand-held captured document, this assumption may be violated, which affects the accuracy of localization (see Fig. 1*a*). We denote this violation as a border visibility (BV) problem. Complicating factors also include the problem of false-positive (FP): the presence of the extended false contrasts in the background (tiles, carpet, blinds, keyboard) (see Fig. 1*b*) and within the document (text lines, photos, PDF417). Despite these shortcomings, there are many publications [6–13] which employ the algorithms that belong to the first class.

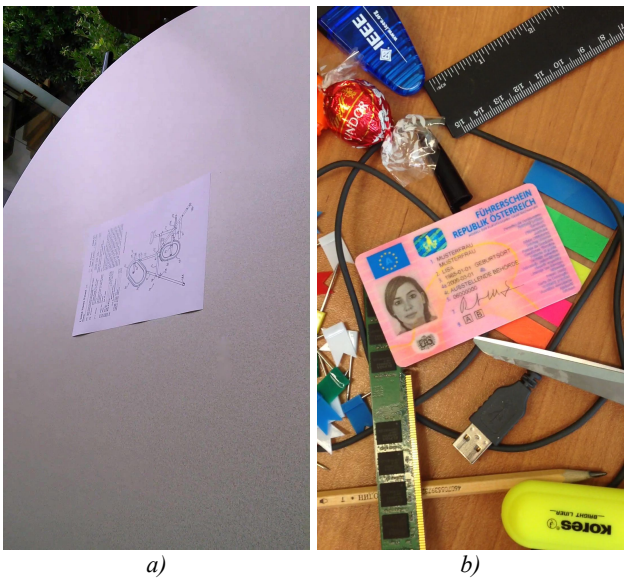

*a)*                            *b)*

*Fig. 1. Example of images with (a) low contrast of one of documents borders and (b) cluttered background. The image (a) is taken from open source SmartDoc dataset and (b) from open source MIDV-500 dataset*

The second class of algorithms is based on salient region detection. The basic assumption in such algorithms is that there is a high contrast between the document and the background. The first approach to the contrast analysis is conventionally referred to as proceeding from the general to the particular (from the model to the document). Attivissimo et al. in [14] employ quadrilateral as a model. They optimize the position of the document maximizing the color difference between the inner and outer area of the quadrilateral. The second approach to contrast analysis can be conventionally referred to as proceeding from the particular to the general: the position of the document is evaluated based on the segmentation into two classes (document, background) image. Ngoc et al. in their work [15] represent the image as an unordered tree of regions, which is obtained by hierarchical segmentation of the image. For segmentation, they use several contrast features at once: the Dahu pseudo distance, the Chi-square distance between histograms of the color sets of pixels. In [16], Leal and Bezerra use similar features to segment the image into two classes. Above mentioned methods of contrast analysis are stable in terms of false rectilinear contrasts present in the image on the background and on the document. However, this method for document localization is computationally expensive.

Neural network methods of image segmentation are also used to improve robustness in the case of complex backgrounds. For example, [17] and [18] employ the original U-net neural network. There are millions of parameters in such networks, so researchers are concentrating on the reduction of the parameters. Ricardo et al. in [19] modified the U-net network architecture, reducing the number of parameters by more than 70%. In [20], the number of parameters was reduced by a factor of 100 via the Fast Hough Transform [21] and its inverse. Still, the employment of neural networks on smartphones remains ambiguous.

The third class of methods is based on document vertices detection [6, 18, 22]. Since there may be abundant false angles in the image, the Region Of Interest (ROI) is employed to improve the stability of the systems. In [18], U-net is employed to estimate ROI, and in [22], a convolutional neural network (CNN) is used. In [6], ROI is not used. Authors search for angles in the entire image, however, the system with angle detection is complementary to the main method based on borders detection.

The majority of the methods proposed in the aforementioned publications do not take into account the constraints of the on-device (smartphone) application. Only in five papers [7–11], the proposed algorithms are designed specifically for smartphones. All of these systems employ Hough-based document localization algorithms. In [7] and in [8], high-frequency structures filtering system is used to overcome the FP problem. In [9], the quadrilaterals are formed by straight lines gained by no filtered edges. Since the ratio of sides is known, not only the standard method with the intersection of four sides is employed, but also the triples of sides are used to form a quadrilateral. This improves the quality of the algorithm output, especially when one document border is not visible. These authors, in another paper [10], suggest ranking quadrilaterals not only via the edge characteristics along the borders but also taking into account the contrast between foreground and background of the quadrilateral. Puybareau and Geraud [11] were also inspired by sali-ence-based methods: prior to search for straight lines, they employ the watershield algorithm to extract contours in the image.

The systems proposed in [7, 8] do not have reference performance evaluations on open datasets. The methods [10, 11] were evaluated on the open dataset SmartDoc [23]. And in spite of the high-quality performance, they do not achieve state-of-the-art scores. The performance of the algorithms proposed in [9] as well as in [10] was evaluated on the open MIDV-500 dataset [24].

In this paper, we propose the on-device document localization system, which also takes a Hough-based approach. We will use the modifications proposed in [9, 10] to make the system more reliable in terms of FP and BV problems. The proposed method allowed for the second best result for SmartDoc dataset in terms of precision, surpassed only by a U-net like neural network. When evaluated on a more challenging in terms of clutter MIDV-500 dataset, the proposed algorithm guaranteed the best precision compared to published methods.

## *1. Background*

This section briefly describes the basic elements of the general pipeline for document localization.

### *1.1 Reconstruction of the document border based on three lines and document's aspect ratio*

In [8], Tropin et al. employed triples of straight lines and the reconstructed side to form a quadrilateral in addition to the standard method which searches for 4 straight lines. To reconstruct the fourth side of the document from the detected three, the aspect ratio of the document and the internal parameters of the camera, the principal point position and the focal length, should be known. To simulate the camera, hereinafter we use a pinhole camera model with a center at $O$, focal length, and known position of the principal point. Let us consider two horizontal lines and one vertical line to be known for the quadrilateral formation. The analytical solution of the fourth quadrilateral border reconstruction is reduced to the calculation of the vertical vanishing point. The authors intersect a known vertical line and the plane which is (i) perpendicular to the horizontal vanishing point and (ii) passing through the center of the camera. The algorithm proposed in [9] is analytical and requires a constant number of calculations.

### *1.2 Quadrilateral filtration according to known geometry properties of the document*

In the pinhole camera model, the projection of a rectangular document on the image plane is generally a quadrilateral (Fig. 2). This transformation is performed via the central projection of a document in the three-dimensional world onto the image plane. The inverse transformation does not preserve the original document's scale. To perform the inverse transformation, the tetrahedral angle $OABCD$ formed by the center of the camera $O$ and the document image $ABCD$ is dissected by the plane $\alpha$. The latter is parallel to the plane $OV_1V_2$, where $V_1$ and $V_2$ are the vanishing points. Next, to get the coordinates of the inverse image, the quadrilateral $ABCD$ is projected onto the plane $\alpha$. If the quadrilateral $ABCD$ is the true image of the document, the inverse image will be a rectangle with the aspect ratio of the document. In the general case, for any convex quadrilateral on the image plane, the inverse image will be a parallelogram.

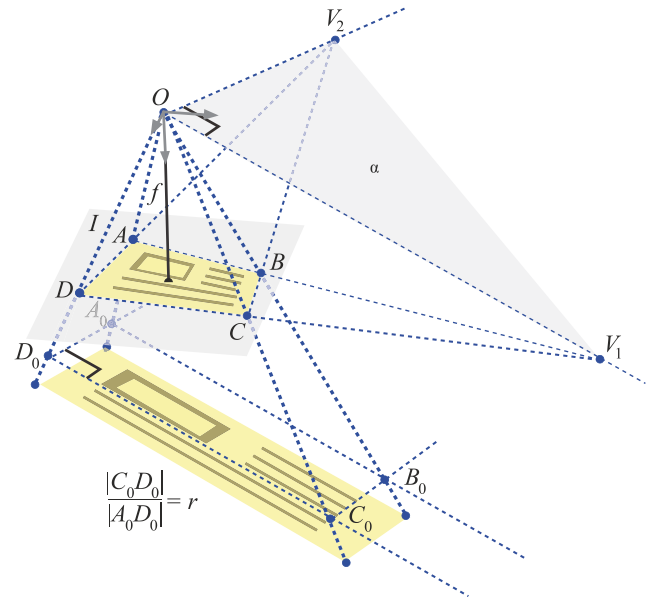

*Fig. 2. Pinhole camera model*

In [9], to filter the quadrilaterals formed by four crossing lines, the difference between a resulting parallelogram and a rectangle with a known aspect ratio ($r$) is used.

### *1.3 Ranking quadrilaterals by contours and contrasts*

To pick the quadrilateral which corresponds to the document among all of the detected quadrilaterals, the latter should be ranked.

The Hough-based methods employ contour ranking of the quadrilaterals. This score is based on the integral characteristics of the contours along the sides $\{b\}$ of the quadrilateral $q$: the intensity of the edges inside $w$ and outside $w'$ of the borders [8], and the shares of non-zero pixels $c$ on the edge map along borders $b$ [5]. In [10], Tropin et al. suggested the following contour score:

$$C = \frac{\sum_{\{b\}} w(b)}{1 + \sum_{\{b\}}(1 - c(b))} - \sum_{\{b\}} w'(b) \qquad (1)$$

Here, $w'(b)$ is the total edge intensity of segments which (i) are located on the same line as $b$, (ii) do not intersect each other, (iii) have one common point with $b$, (iv) are 10 pixels long. The computational complexity of such ranking for a single quadrilateral is $O(1)$, provided that there are pre-computed arrays of integral statistics along the lines of the borders of the quadrilateral $q$.

To improve the performance on complex backgrounds (Fig. 3*a*), the authors of [10] employed the contrast estimation of the quadrilateral candidates. It is based on the $\chi^2$ distance between the RGB pixel histograms of the inner and outer areas of the quadrilateral. In order for the contrast score to be independent of the size of the quadrilateral candidates, $\chi^2$ distance is calculated on a projectively normalized images (Fig. 3*bc*).

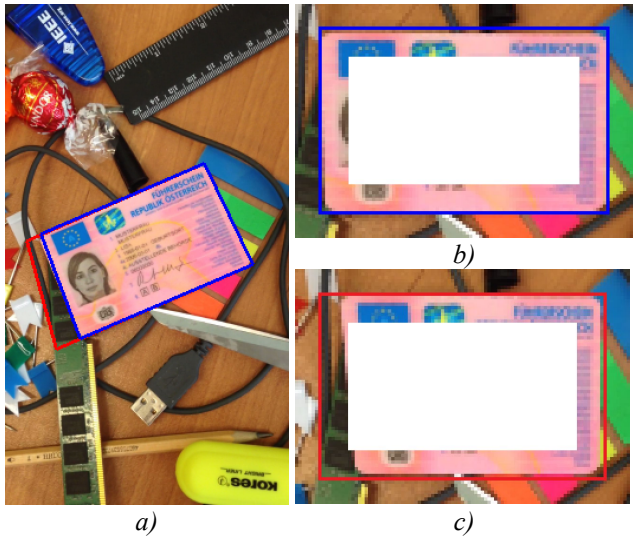

*Fig. 3. Example of situation when a true quadrilateral, marked by blue color on (a), has lower contour score (since its left side is not precise) than a contour score of a false quadrilateral (red color): the contour score of the true quad is about 9063 vs 10919 – false quad. However, the contrast score of the true quadrilateral is higher than contrast score of the false quadrilateral 170 vs 143. The figures (b) and (c) contain an outer and an inner region of the true and false quads correspondingly (white rectangle in the center restricts the inner region). These regions are used for the contrast score estimation*

Since the contrast computation for each quadrilateral individually is computationally expensive, in [10] suggested dividing the ranking of the quadrilateral candidates into two steps: (1) contour score, to drop the alternatives which are lower than $K$ th place in the ranked list, (2) a linear combination of contour and contrast scores for the remaining alternatives (the final ranking of the quads on the Fig. 3*a* was based on linear combination of the contrast and contour scores where the latter was multiplied by 0.011).

## 2. *The proposed algorithm*

To reduce the image noise, the original image is compressed isotropically so that its new shortest side is 240 px. We will consider main steps of the proposed algorithm on the image illustrated on the Figure 4*a*.

Since, according to the problem statement, the document borders are two pairs primarily horizontal and primarily vertical lines, the search for edges and lines is divided into two non-overlapping processes. Let us begin by describing the detection of horizontal borders (the search for vertical borders is identical).

### 2.1 *Edge detection*

Borders detection begins with channel-by-channel morphological filtering of the image. First, an opening operation is applied with a window wing of 1 pixel, and then a closing operation is applied with the same window. Thus, the filtering of valleys (local minima) and ridges (local maxima) with a width of two pixels is performed. After the morphological filtering of local minima and maxima within two pixels, the image with the derivative along the Y-axis is calculated (the kernel filter is (1;-1)). Then, for each pixel, the derivative values are averaged over each of the RGB channels. After that, a non-maxima suppression operation is performed inside each column (a structural element with a width of 1 pixel) for the pixels with absolute derivative values greater than 1. Then, the horizontal connectivity components are collected: for each pixel, three pixels on the left and three on the right are considered its neighbors. These components are filtered by size: all of the components less than 10% of the minimum between (i) the size of the maximum component and (ii) half the image width are dropped. The edge map is filled with the same constant values at points covered by the remaining components. Finally, the edges are blurred via the Gaussian function. The morphological operations at the first stage and Gaussian filtering at the last stage are performed for each column independently.

The search for vertical edges is performed in the same way as the search for horizontal edges. The image with horizontal and vertical edges is illustrated on the Figure 4*b*.

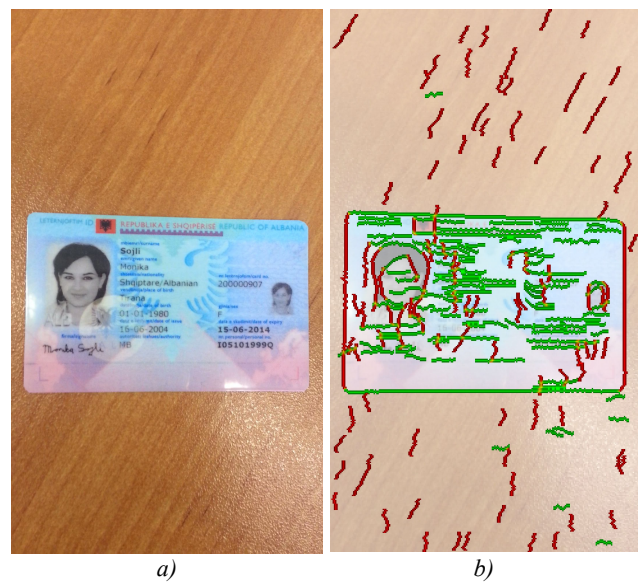

*Fig. 4. Input image (a) and edge maps (b): red color correspond to primarily vertical edges, green one – to primarily horizontal edges*

### *2.2 Line detection*

After two edge maps are computed, a straight line search is performed. This step yields two sets of lines: primarily horizontal and primarily vertical. Without loss of generality, let the height of the image be greater than its width. Then, to search for the primarily vertical lines, the image is divided into three roughly equal non-overlapping parts. For each part, the Fast Hough Transform (FHT) [21] is performed and the value of the global maximum among all three parts is determined. After that, 15 local maxima (the maximum is not strict in all directions except for two directions: up and left) are sequentially selected in each part, provided that (i) the current maximum exceeds the 20% threshold of the global maximum and (ii) the current maximum lies more than 10 pixels by $l_2$ norm away from the one which has been already selected (Fig. 5). Then, using the inverse FHT, the selected maxima are converted into straight lines in the image. Thus, 45 vertical lines are detected. The horizontal lines are detected in the same way as the vertical ones, but the FHT is performed for the entire image and only 15 maxima are selected, resulting in 15 lines (Fig. 6).

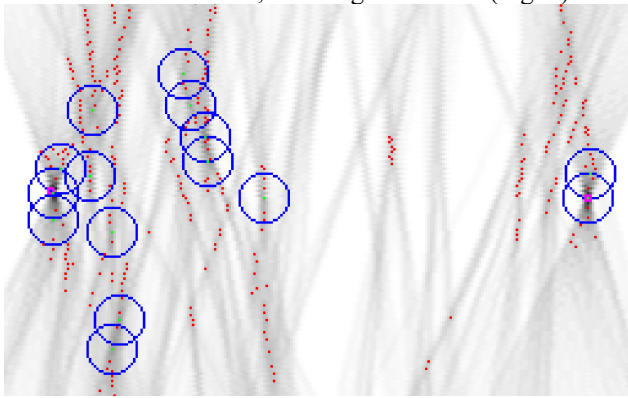

*Fig. 5. Inverted FHT image of a middle part of the vertical edge map with local maximums (red dots), 15 local maximums which were selected (green dots) and their neighborhoods (blue circles). The images of true vertical sides of the card are highlighted by magenta squares*

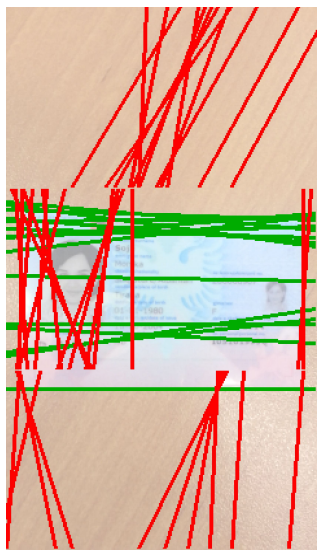

*Fig. 6. Red segments mark primarily vertical lines detected on corresponding parts of the vertical edge map; green segments visualize primarily horizontal lines*

### *2.3 Two-step ranking system*

We implemented a two-step ranking of the contour and contrast characteristics of the quadrilaterals (see sec. 1.3). At the first step, the contour score (see eq. 1) is calculated for all of the generated quadrilaterals which were pre-filtered (this is explained further). At the second step, only for each of the $K = 4$ quadrilateral candidates which have the maximum contour score, a linear combination of contour and contrast score is computed. The candidate with the maximum combination of scores is selected for the next step described in section 2.4.

#### *2.3.1 Quadrilateral formation and filtering*

The first step filters out and sorts the quadrilaterals formed by both quadruples and triples (see sec. 1.1) of straight lines. Such formation is a result of a consecutive brute-force search: by four lines and by three lines. And since only $K = 4$ candidates are required as the result of the first ranking step (see sec. 2.3), a maximum of 4 candidate quadrilaterals should be formed by each search.

Based on the constraint of 4 candidates, we maintain a heap for a minimum of 4 elements when searching for the quadrilaterals. The contour score (1) of a quadrilateral includes the reward (total intensity along the four sides $w$) and the penalty based on the fraction of non-zero pixels $c$ and the intensity of the edges $w'$ outside of the quadrilateral. If the value of the reward of the tested quadrilateral is less than that of the root of the heap, then it makes no sense to consider this candidate further and it may be rejected. We employ such filtration in each search.

When calculating the contour score, all of its components $(w, c, w')$ are compared to the minimum and maximum thresholds; if these components are not within the allowed range, the candidate is filtered out.

To filter out the quadrilaterals unavailable in the centrally projective model (in our experiments camera's back focal length $f$ equals to 0.705 from a diagonal of the image in working resolution and a principle point is in the center of the image), we reconstructed an inverse image of the document. The shape of this image is a parallelogram (see sec. 1.2). If the difference between the aspect ratio of the parallelogram and the ideal aspect ratio exceeds 7% of the ideal aspect ratio or if the difference between the angle of the parallelogram and 90 degrees is more than 5 degrees, then the candidate with such an inverse image may be filtered out. This filtration is used only in quadrilateral search by four lines. During the search by three lines, this filtration is useless because all of the quadrilaterals already correspond to the central projective model of the rectangle with a known aspect ratio.

If the candidate passed through all of the filtering stages, and it was not rejected at any point, its contour score is calculated. If the contour score is greater than the value at the root of the heap, it is added to the heap while maintaining the size of the latter.

The top $K$ alternatives from a brute-force search by four lines are illustrated in the first row of the Figure 7, the top $K$ alternatives by three lines – in the second row of the same figure.

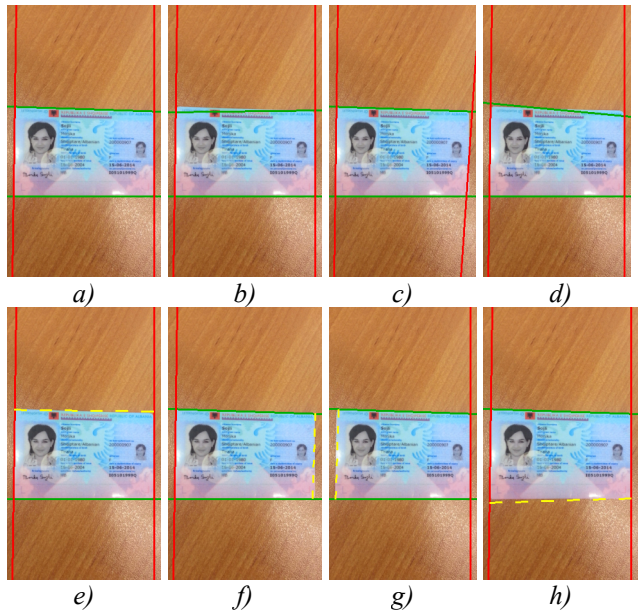

*Fig. 7. Top 4 quadrilateral candidates by contour score formed by 4 lines are illustrated in the first row, top 4 quadrilateral candidates formed by 3 lines – in the second row. Lines that were used for quadrilateral formation have red/green color depending on orientation, restored borders are yellow dotted segments*

### 2.3.2 *Quadrilateral reordering*

On the second step of the ranking only $K$ out of $2K$ quadrilaterals with max contour score are reordered based on the combination of the contour and contrast scores. In case of quadrilaterals illustrated on the Figure 7, the top $K$ quadrilaterals ordered by decreasing of the contour score are the following: *a, e, f, b*.

For each of them contrast score based on $\chi^2$ distance (see sec. 1.3) is calculated and linearly combined with the contour score. Since in this particular case the ordering by contrast coincides with the ordering by the contour score, and since the combination coefficient used in our system is positive, the answer after the second step of ranking is remains the same. This quadrilateral candidate is shown on the Fig. 7*a*.

### 2.4 *Quadrilateral refining*

Since the working resolution $(240 \times 426)$ is 4.5 times smaller than the original resolution, to increase the accuracy of the document localization in the original resolution, we implemented the refining of the detected document borders at a scale 3 times the working resolution. We consider a vicinity of 2 pixels (in the working resolution coordinate system) along each border. We employ the similarity transformation to each of the resulting regions to make them (i) horizontal and (ii) scaled 3 times the working resolution. Then we apply the edge detection module described in the 2.1 section, calculate the FHT for the entire image, and evaluate the global maximum. The refined position of each border is calculated by the inverse FHT of the corresponding maximum at the FHT image. A similar quadrilateral refinement stage was used in [6, 18].

### 3. Experiments

We tested our algorithm on two open datasets of smartphone camera-captured images: SmartDoc [23] and MIDV-500 [24].

### 3.1 *Datasets*

The first dataset, SmartDoc, includes about 25 000 images of A4 page. For a document in each image, there is a ground-truth quadrilateral obtained in semi-automatic mode [25]. The dataset is divided into five parts, but we are interested only in the first four. Only the latter include the images of a single document with a given aspect ratio. SmartDoc's fifth part includes the images of the target document (A4 page) placed on top of a stack of papers. Thus, in addition to the target document borders, four sides of another A4 page are partially visible in the image. It is impossible to distinguish the target document from the one below it solely based on the borders, i.e. without information about the target document content.

The second dataset, MIDV-500, includes 15 000 images of identity documents (50 different types of documents). The dataset contains images captured under several conditions: different background (table, keyboard, clutter), a document is either placed on the surface or held in hands, a document is either partially outside of the frame or within the frame completely. Regardless of whether the document image is completely within the frame or not, there is ground truth: a ground-truth document quadrilateral (even for documents which are completely out of frame) for each image in the dataset. The ground-truth quadrilaterals are marked manually. In addition to the ground-truth, the true aspect ratio of each document is known.

### 3.2 *Problems and corresponding metrics*

To evaluate the system's performance when localizing the quadrilateral of the outer boundaries of the document, the metrics of quality based on the maximum discrepancy between the vertices are used [8, 9]. The index introduced by Tropin et al. in [9] is the best way to evaluate the system in our case. This index is based on the maximum discrepancy between the corresponding vertices of the resulting quadrilateral of the system and the ground-truth quadrilateral in the coordinate system of the resulting quadrilateral of the system. Let $q$ be a resulting quadrilateral, $m$ be a ground-truth quadrilateral, and $t$ be a rec-tangle with linear sizes of the document template. Let $H$ represent a homography such that $Hq = t$, so

$$D(q,m,t) = \max_i \frac{\|t_i - Hm_i\|_2}{P(t)}, \qquad (2)$$

where $P(t)$ is a perimeter of the template. This index would be upgraded [26] if we had to maximize not the quality of the quadrilateral detection but the quality of text recognition inside the region-of-interest of the document.

Since the quadrilateral of the document should be defined precisely up to the vertices renumbering, the final statistics based on (2) is as follows:

$$\text{MinD}(q,m,t) = \min_{q^{(i)} \in Q} D(q^{(i)}, m, t), \qquad (3)$$

where $Q = \{[a,b,c,d];[b,c,d,a];[c,d,a,b];[d,a,b,c]\}$ is a set of quadrilaterals gained by renumbering of vertices of $q$.

$$\text{MeanIoU}(q,m,I) = \frac{1}{2}\left(\frac{\text{cnt}_1(I_q \cap I_m)}{\text{cnt}_1(I_q \cup I_m)} + \frac{\text{cnt}_1(\overline{I_q} \cap \overline{I_m})}{\text{cnt}_1(\overline{I_q} \cup \overline{I_m})}\right),$$

The IoU (Intersection over Union) metric is used in the case of the segmentation into two classes (background, foreground):

$$\text{IoU}(q,m) = \frac{\text{area}(q \cap m)}{\text{area}(q \cup m)}. \qquad (4)$$

This metric is more universal because it can be used to evaluate both the quality of the segmentation and the quality of the detection of the quadrilateral formed by the outer borders of the document.

According to the SmartDoc competition protocol, the quality metric (4) must be calculated in the coordinate system of the ideal quadrilateral.

$$\text{IoU}^{gt}(q,m,t) = \frac{\text{area}(Mq \cap t)}{\text{area}(Mq \cup t)}, \qquad (5)$$

where $M$ is a homography such that $Mm = t$.

We denote the second variation of the metric (4) as MeanIoU:

$$\qquad (6)$$

where $I_q$ and $I_m$ are corresponding binary masks of $q$ and $m$ with the same size as the input image $I$, $\text{cnt}_1$ is a count of true pixels in the image.

The quality metrics described above were implemented in Python 3. The code can be found in the supplementary materials.

### 3.3 Results of experiments on SmartDoc

Let us consider the Tab. 1. It illustrates the evaluation of already available systems according to the averaged index $\text{IoU}^{gt}$ (5). Each row of the table corresponds to a system. The systems are listed in descending order by the average accuracy for all four SmartDoc subsets. The results clearly show that based on the overall accuracy, our system is the second best, surpassed only by [19] proposed by das Neves et al. This method is based on a U-net like neural network.

On the Fig. 8 we illustrated the result of proposed algorithm on images from SmartDoc. The image with the worst result of the algorithm according to the quality metric (5) is shown on the Fig. 8*a*. The images Fig. 8*b-d* corresponding to quantile 1–3. The image with the highest result is illustrated on the Fig. 8*e*.

*Table 1. Comparison with the-state-of-the-art systems for the SmartDoc dataset*

| System | Bgr. 1 | Bgr. 2 | Bgr. 3 | Bgr. 4 | All | Run-time (ms) / Device |
|---|---|---|---|---|---|---|
| das Neves et al. [19] | **[0.9912; 0.9934]** | | | | | 59 / Intel Core i7 8700, 8 GB RAM, 6 GB NVIDIA GTX 1060 |
| Our | 0.9886 | 0.9858 | 0.9896 | 0.9806 | 0.9866 | 65 / Apple A8 (iPhone 6) 1 GB RAM |
| Zhukovsky et al. [6] | 0.9885 | 0.9833 | 0.9897 | 0.9785 | 0.9856 | – / – |
| Zhu et al. [18] | 0.9876 | 0.9839 | 0.9830 | 0.9843 | 0.9848 | 564 / Inter Xeon(R), 2CPU E5-2603 v4, 4GPU Nvidia-1080 |
| Ngoc et al. [15] | 0.9850 | 0.9820 | 0.9870 | 0.9800 | 0.9838 | 3700 / Intel i7 CPU, 8 GB RAM |
| Tropin et al. [10] | 0.9830 | 0.9740 | 0.9830 | 0.9700 | 0.9787 | – / – |
| Javed and Shafait [22] | 0.9832 | 0.9724 | 0.9830 | 0.9695 | 0.9777 | 320 / Intel i5-4200U CPU, 8 GB RAM |
| Leal and Bezerra [16] | 0.9605 | 0.9444 | 0.9647 | 0.9300 | 0.9516 | 430 / Intel(R) Core (TM), i5-3570K CPU, 16 GB RAM |
| Puybareau and Geraud [11] | 0.9050 | 0.9360 | 0.8590 | 0.9030 | 0.9007 | 40 / – |

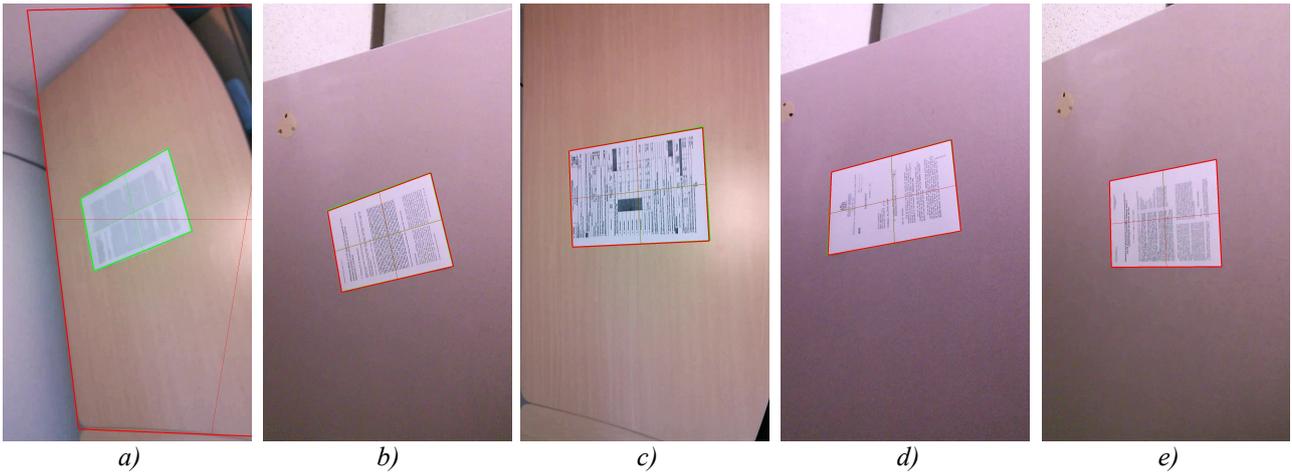

*Fig. 8. The example of the output quadrilateral (red color) and ground-truth quadrilaterals (green color) on the images from SmartDoc. The images are selected as follows: all images from SmartDoc (backgrounds 1-4) with non-empty result (for visualization purpose) are sorted by non-decreasing of index (5) and then 0% (a), 25% (b), 50% (c), 75% (d), 100% (e) quantiles are picked*

### 3.4 Results of experiments on MIDV-500

MIDV-500 dataset does not offer a standard protocol for measuring the accuracy of document localization systems: there are different quality metrics as well as different test subsets of MIDV-500. The use of different metrics is justified, for example, when the objectives differ (see sec. 3.2 for more details).

Let us consider the table 2. The first four columns represent the accuracy measurements on MIDV-500 for 4 document localization systems. All of these systems, except for [26], do not take into account the document content. A brief description of the MIDV-500 subsets is given in the third column of the table; a detailed description, as well as the subsets themselves, can be found in the supplementary materials. The fifth column represents the measurements of the accuracy of our system.

The Fig 9*a-e* contains images from MIDV-500. They are selected by analogy with Fig 8.

*Table 2. Comparison with the state-of-the-art systems for the MIDV-500 dataset*

| System | Statistics | Set | Accuracy | Our |
|---|---|---|---|---|
| Tropin et al. [10] | Averaged IoU$^{gt}$ | Full (15000 images) | 0.8700 | **0.9056** |
| | | At least 3 vertices are within the frame (11965) | 0.9610 | **0.9788** |
| | | All 4 vertices are within the frame (9791) | 0.9720 | **0.9830** |
| Tropin et al. [9] | The percentage of images: MinD $\leq$ 0.017 | Full (15000) | 80.50% | **82.43%** |
| | | At least 3 vertices are within the frame (11991) | 92.66% | **94.92%** |
| | | All 4 vertices are within the frame (9791) | 94.66% | **96.77%** |
| Sheshkus et al. [20] | Averaged MeanIoU | At least 3 vertices are within the frame (11965) | 0.9600 | **0.9862** |
| Guillaume et al. [27] | The percentage of images: IoU $\geq$ 0.9 | The percentage of the document area inside the frame is higher or equal than 90% (13586) | **97.00%** | 93.99% |

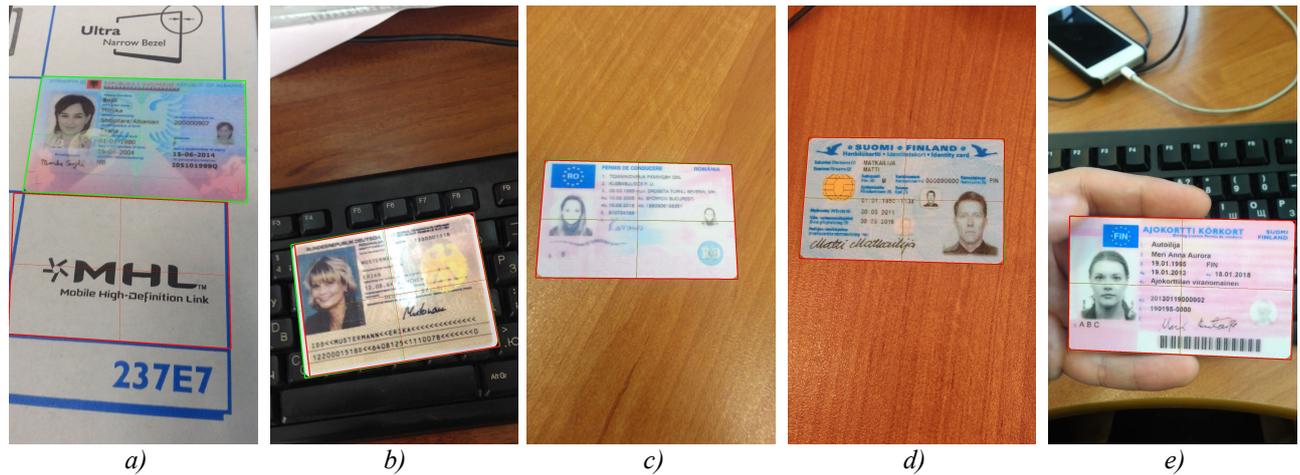

*Fig. 9. The example of the output quadrilateral (red color) and ground-truth quadrilaterals (green color) on the images from MIDV-500. The images are selected as follows: all images from MIDV-500 with non-empty result are sorted by non-decreasing order of index (5) and then 0% quantile (a), 25% (b), 50% (c), 75% (d), 100% (e) are picked*

### 3.5 *Run-time measurement*

For run-time measurement, an iPhone 6 with Apple A8 CPU and 1 GB RAM in single treading mode was used. We tested our system twice: on 100 random images from SmartDoc and on 100 random images from MIDV-500. The running time per frame for a subset of SmartDoc images is 65 ms, and 102 ms for MIDV-500. The difference in running time between these two subsets is explained by the overall lower number of false straight-line contrasts in the SmartDoc dataset compared to the MIDV-500 dataset.

### 4. Discussion

Apart from das Neves et al, the proposed algorithm yields (Tab. 1) one of the highest accuracy values for the first, the second and the third backgrounds and a negligible loss in the case of the fourth background of the SmartDoc dataset. The [18] method, which allows for great accuracy on the fourth background, also uses the U-net network to preliminary estimate the ROI of the document image. Our method outperforms the Hough-based systems [10, 11] specifically designed to work under smartphone hardware.

The accuracy values in Tab. 2 clearly demonstrate that the proposed method outperforms other systems which do not take into account the document content. The system proposed in [27] outperforms our system and generate the baseline for a document outer border detection with the knowledge of the inner structure of the document.

In order to further fix the baseline performance on MIDV-500, in addition to publishing the quality metrics and subsets, we also illustrated the resulting quadrilaterals (the output of our system) and provided a script for statistics calculations in the supplementary materials.

The last column of the Tab. 1 illustrates the measurements of the document localization time per frame as well as the parameters of the device. The characteristics of the devices are different, as well as the running time. However, this data shows that (i) the running time of our system is short and (ii) unlike other systems, our algorithm runs on a mobile processor, which is inferior to other devices in terms of performance.

Our algorithm is comparable to the systems proposed in [9, 10]: they use the same device and CPU mode when measuring the running time for the MIDV-500 subset. The list of methods ordered by the execution time is as follows: 67 ms/frame [9], 88 ms/frame [10], 102 ms/frame for the proposed system.

Proposed algorithm afford to cope with cases when BV and FP problems occur (Fig. 10*a* and 10*b* correspondingly).

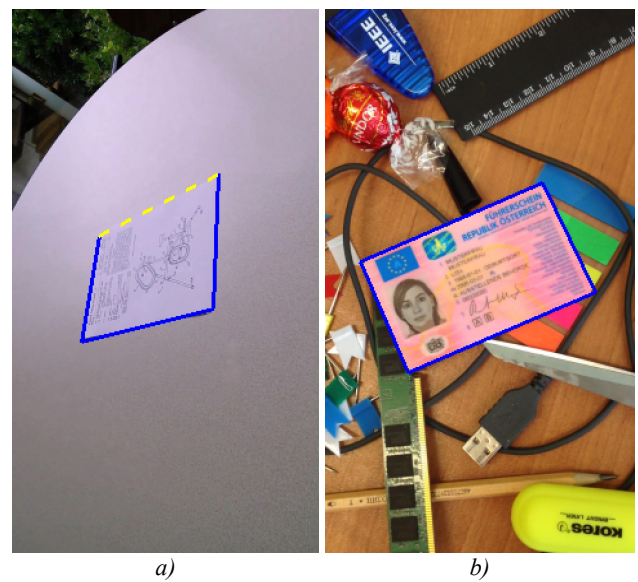

*a)*    *b)*

*Fig. 10. Example of resulting quadrilaterals on images with challenging conditions. Blue sides of the quadrilateral found by line detection module, yellow dotted side is restored by three other borders and known aspect ratio.*

Based on the Fig. 8 and Fig. 9 we conclude that in at least 75% of cases the error of proposed system is small. This thesis can be refined if you pay attention to the 5th line of the Tab. 2: in 94% the error is small.

## Conclusion

This paper describes a new algorithm for the detection of the quadrilateral formed by the outer borders of a document. There were two main requirements for the developed algorithm: applicability to the case when there is no a priori information about the document content, and sufficient performance to operate on smartphones in real time. We chose the Hough-based approach, which turned out to be the only working option under the aforementioned constraints. Algorithms of this class analyze the visible quadrilateral outline of the document. The former are characterized by low computational complexity, which allows for the fulfillment of both requirements. However, all of the previously known Hough-based solutions have a significant disadvantage: in terms of accuracy, they are inferior to the algorithms which solve the same problem while there are no computational constraints.

Here, we proposed the Hough-based algorithm which in addition to the standard techniques employs geometric invariants of rectangular projections as well as color characteristics of bordering regions. Due to the first advancement, it was possible to significantly increase the robustness of the system when one of the document borders is not visible. The second one improved the accuracy of the system on images with a large number of edges. Combined with other more standard optimizations, such as multiscale analysis, the developed approach, when tested on two open relevant datasets, achieved the accuracy typical for algorithms without any performance limitations.

For the SmartDoc dataset (printed documents with arbitrary content), the proposed method is the second best, demonstrating the average accuracy of around 0.985 in the $IoU^{gt}$ (5) index with an average execution time of 65 ms per frame for the iPhone 6 in the single-threaded mode of operation. For the MIDV-500 dataset (identity documents), the proposed method is the best in terms of accuracy: the average corresponding values are 0.98, 100 ms on a subset of the data with a large number of false straight-line contrasts and the captured document being out of the frame. The average execution time confirms the applicability of the proposed algorithm to the on-device recognition systems. Let us emphasize that in both cases, we compared our method with all known algorithms tested on the specified datasets, regardless of their performance.

Achieving the highest accuracy on the MIDV-500 dataset is an important result. This dataset contains images captured in the close-up "hand-held" mode, which simulates the use-case of real-time recognition. At the same time, we cannot say that this dataset is "simpler" than SmartDoc. It is known [10] that good quality on SmartDoc does not guarantee high quality on MIDV-500. We hope that the publication of these baselines for the MIDV-500 dataset for document localization algorithms, which do not take into account information about document content, will allow for a better comparison of the algorithms of this class in the future. The results of document localization of proposed system as well as an implementation of used quality metrics are available for download by the link https://github.com/SmartEngines/hough_document_localization.

To conclude, we note that, apparently, the achieved quality is not the Hough-based methods limit. Part of the errors for the datasets are caused by errors of the greedy selection of maxima for the Hough image, thus the peak corresponding to the true straight line can be lost. This is a well-known problem of Hough-analysis. There are various methods, including very high-performance ones, which address this issue. These methods can be employed for the further development of the proposed approach.

*Authors' information*

**Daniil Vyacheslavovich Tropin** (b. 1995), graduated from Moscow Institute of Physics and Technology (State University) in 2019, majoring in Applied Mathematics and Informatics. Currently he is a postgraduate student at the latter and lead programmer at FRC CSC RAS. Research interests are image processing and computer vision. E-mail: daniil_tropin@smartengines.com.

**Alexandr Mikhailovich Ershov** (b. 1999), student of Lomonosov Moscow State University at the chair of functional analysis. He is currently working as a programmer in Smart Engines. Research interests are banach space geometry and programming. E-mail: a.ershov@smartengines.com

**Dmitry Petrovich Nikolaev** (b. 1978), Ph. D. in Physics and Mathematics, is a head of the laboratory at the IITP RAS, a technical director of Smart Engines Service LLC. He graduated from Lomonosov MSU in 2000. His major research interests include machine vision, algorithms for fast image processing, pattern recognition. E-mail: dimonstr@iitp.ru .

**Vladimir Viktorovich Arlazarov** (b. 1976), PhD, graduated from Moscow Institute of Steel and Allows in 1999, majoring in Applied Mathematics. Currently he works as head of division 93 at FRC CSC RAS. Research interests are pattern recognition and machine learning. E-mail: vva@smartengines.com .

*The mailing address (example, not published):*

*Mob. tel. of Daniil Tropin*